\documentclass{article}
\usepackage{spconf,amsmath,graphicx}
\usepackage{xcolor}
\usepackage{adjustbox}
\usepackage{tabularx}
\usepackage{ragged2e}
\usepackage{booktabs}
\usepackage{multirow}
\usepackage{hyperref}

\usepackage{subcaption}

\makeatletter
\def\@maketitle{\newpage
 \null
 \vskip 2em
 \begin{center}
  {\large \bf \@title \par}
  \vskip 1.5em
  {\large \lineskip .5em
  \begin{tabular}[t]{c}
    {\em Junkai Wu$^{\sharp *}$\thanks{$^*$ Equal Contribution.}, Xulin Fan$^{\flat *}$, Bo-Ru Lu$^\sharp$, Xilin Jiang$^\natural$} \\
    {\em Nima Mesgarani$^\natural$, Mark Hasegawa-Johnson$^\flat$, Mari Ostendorf$^\sharp$}
  \end{tabular}\par}
  \vskip 0.75em
  {\large \lineskip .5em
  \begin{tabular}[t]{c}
    $^\sharp$University of Washington, $^\flat$University of Illinois Urbana-Champaign, $^\natural$Columbia University
  \end{tabular}\par}
 \end{center}
 \par
 \vskip 1.5em
}
\makeatother

\title{Just ASR + LLM? A Study on Speech Large Language Models' Ability to Identify and Understand Speaker in Spoken Dialogue}

\name{Junkai Wu$^{\sharp *}$\thanks{$^*$ Equal Contribution.}, Xulin Fan$^{\flat *}$, Bo-Ru Lu$^\sharp$, Xilin Jiang$^\natural$, Nima Mesgarani$^\natural$, Mark Hasegawa-Johnson$^\flat$, Mari Ostendorf$^\sharp$}

\address{$^\sharp$University of Washington, $^\flat$University of Illinois Urbana-Champaign, $^\natural$Columbia University}

\begin{document}
\maketitle

\begin{abstract}
In recent years, we have observed a rapid advancement in speech language models (SpeechLLMs), catching up with humans' listening and reasoning abilities.  SpeechLLMs have demonstrated impressive spoken dialog question-answering (SQA) performance in benchmarks like Gaokao, the English listening test of the college entrance exam in China, which seemingly requires understanding both the spoken content and voice characteristics of speakers in a conversation. However, after carefully examining Gaokao's questions, we find the correct answers to many questions can be inferred from the conversation transcript alone, i.e.\ without speaker segmentation and identification. Our evaluation of state-of-the-art models Qwen-Audio and WavLLM on both Gaokao and our proposed "What Do You Like?" dataset shows a significantly higher accuracy in these context-based questions than in identity-critical questions, which can only be answered reliably with correct speaker identification. The results and analysis suggest that when solving SQA, the current SpeechLLMs exhibit limited speaker awareness from the audio and behave similarly to an LLM reasoning from the conversation transcription without sound. We propose that tasks focused on identity-critical questions could offer a more accurate evaluation framework of SpeechLLMs in SQA.
\end{abstract}

\section{Introduction}
\label{sec:intro}

Speech large language models (SpeechLLMs), trained with thousands or more hours of data for various questions, have demonstrated state-of-the-art performance in many speech tasks, including automatic speech recognition, speech translation, and emotion recognition. One of the tasks that SpeechLLMs are believed to solve reliably is spoken dialogue understanding. In spoken dialogue understanding, two or more speakers are talking to each other, and SpeechLLMs are tested on \textit{what is said} and \textit{who said what}. To answer these questions correctly, SpeechLLMs are expected to comprehend both the semantic content of the conversation and the acoustic characteristics of speakers. The high performance of SpeechLLMs in SQA benchmarks like Gaokao suggests that SpeechLLMs possess both of these capabilities.

\begin{figure*}[ht!]\captionsetup[subfigure]{font=}
    \includegraphics[width=\linewidth]{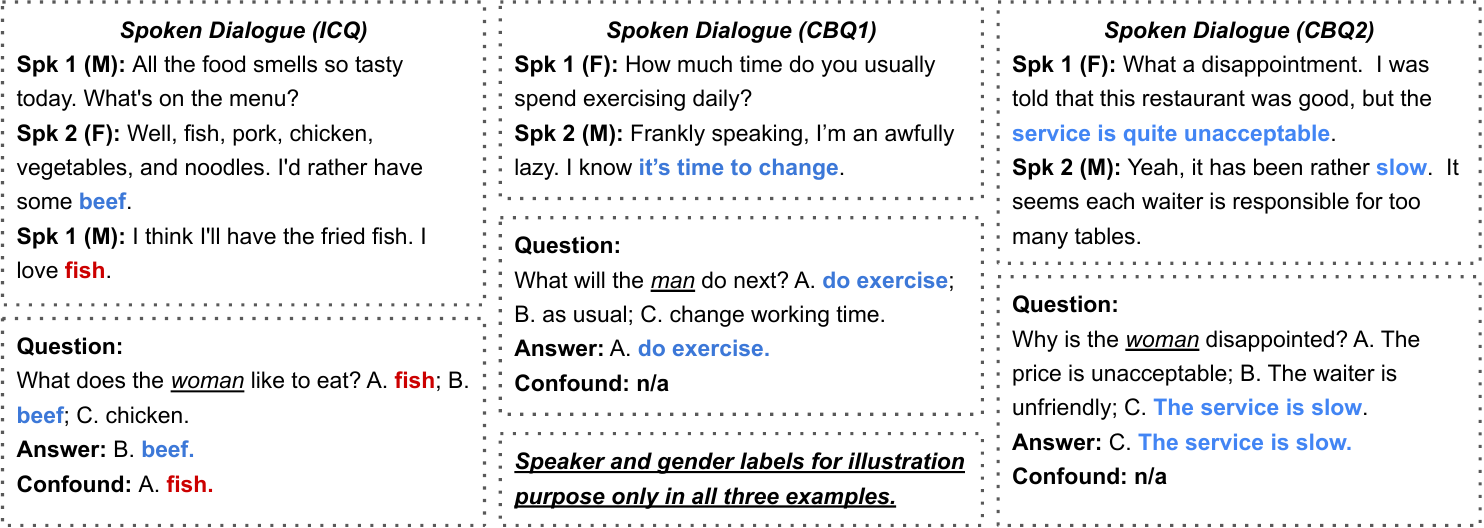}
    \caption{Examples of an identity-critical question (ICQ) and two context-based questions (CBQs). The male (M) and female (F) indicators in the transcript are only included in oracle experiments.} 
  \label{fig:split}
\end{figure*}

However, upon careful inspection, we found SpeechLLMs frequently made mistakes on some surprisingly simple questions. We hypothesize that SpeechLLMs are worse at differentiating speakers' voices than commonly assumed. To test our hypothesis, we separate out speaker-related SQA questions that require identifying a particular speaker's voice characteristics as \textbf{Identity-Critical Questions} (ICQs) and the remaining questions that ask for conversation content solvable without identifying any speaker as \textbf{Context-Based Questions} (CBQs). We evaluate the performance of state-of-the-art SpeechLLMs Qwen-Audio \cite{qwenaudio} and WavLLM \cite{wavllm}, as well as cascaded ASR + LLM systems unaware of any speaker acoustics, on both types of questions in the popular Gaokao benchmark and our proposed benchmark ``What Do You Like?". For the latter, we control the speakers, the content of the conversations, and the questions asked to investigate the behaviors of SpeechLLMs better. Results show that both SpeechLLMs and the ASR + LLM system perform significantly worse on ICQs than CBQs in both benchmarks. Our analysis suggests that current SpeechLLMs perform similarly to ASR + LLM: they successfully answer CBQs with the content of the conversation and fail on ICQs, unable to differentiate speakers' characteristics such as gender.

We hope our observations can inspire a reconsideration of current SpeechLLMs' SQA ability and promote future development of stronger SpeechLLMs and more comprehensive SQA benchmarks.

\section{Related Works}
\label{sec:related_works}
\subsection{SpeechLLMs}
Speech large language models (SpeechLLMs) have received much attention since the success of text-only language models. LTU-AS \cite{gong2023ltuas}, built upon LTU \cite{gong2023ltu}, features a Whisper encoder \cite{whisper} and a LLaMA language model \cite{touvron2023llama1}. After finetuning with low-rank adaptation (LoRA; \cite{hu2021lora}) on a speech and audio question-answering dataset with 9.6M (audio, question, answer) tuples, the model can perceive various speech and audio events and shows satisfactory results on a wide range of speech and audio-related tasks like audio classification, speech recognition, and emotion recognition. Qwen-audio \cite{qwenaudio} uses a Whisper+Qwen \cite{bai2023qwentechnicalreport} architecture and a Whisper-style multi-task training paradigm. WavLLM \cite{wavllm} utilizes a Whisper encoder and a WavLM \cite{wavlm} to encode semantic and acoustic information respectively. A two-stage curriculum training approach is applied to augment the generalization ability of the language model.

\subsection{Speech and Audio Understanding Benchmarks}
Speech understanding benchmarks like SUPERB \cite{superb} and SLUE \cite{slue1, slue2, arora-etal-2024-evaluation} span tasks such as emotion recognition, sentiment analysis, and question answering, primarily evaluating self-supervised models and ASR systems. Several recently proposed benchmarks focus on evaluating SpeechLLMs. 
SD-Eval \cite{ao2024sdeval} is a dialogue understanding benchmark focusing on four aspects: emotion, accent, environment, and age. Evaluation on SD-Eval shows that SpeechLLMs outperform cascaded ASR+LLM systems across all four aspects, but have a significant gap below the LLMs with ground-truth transcripts
and paralinguistic/environmental labels. AudioBench \cite{wang2024audiobenchuniversalbenchmarkaudio} finds cascaded ASR+LLM systems outperform SpeechLLMs for tasks like ASR and speech question-answering, while SpeechLLMs excel when the tasks involve non-speech sound or paralinguistic information, like audio captioning and gender recognition. It is worth noting that in AudioBench, speech question-answering is identified as a speech-intensive task where the cascaded system performs better than all the SpeechLLMs tested including WavLLM, Qwen-audio, and SALMONN \cite{salmonn}. This may result from SpeechLLMs' over-reliance on querying the text information on SQA samples that require paralinguistic information.



%
%

\section{Identity-Critical and Context-Based SQA Questions}
\label{sec:issue_demo}

Based on our preliminary experiments with SpeechLLMs, we observed failure cases when SpeechLLMs are confused about ``who said what.'' We hypothesize that SpeechLLMs may not fully leverage the speaker ``who'' information in the audio to answer these questions, since the ``what'' part is usually mentioned by another speaker but mis-selected by SpeechLLMs. To analyze the speaker factor alone and eliminate other factors, we propose a categorization of questions based on whether they can be answered correctly with or without correct speaker identification from the audio. In this section, we describe identity-critical questions (ICQs) and context-based questions (CBQs), then introduce an automatic method to classify each question as an ICQ or a CBQ.

\subsection{Identity-Critical Question (ICQ)} 
\label{sec:icq}
An ICQ demands an accurate identification of the speaker referenced in the question to answer correctly. When multiple speakers' preferences/actions/events are mentioned, the model must capture ``\textit{who said what}'' and leverage the target speaker's information and linguistic content to provide the correct answer. In the ICQ example in \autoref{fig:split}, the question requires the model to both recognize and differentiate the speaker preferences (``fish'' and ``beef'') from the content by their voices. In ignorance of speaker information in the audio, SpeechLLMs usually randomly output an answer from the two candidates mentioned in the audio.

\subsection{Context-Based Questions (CBQ)}
\label{sec:cbq}
A CBQ does not require the identity of the speaker referenced in the question. This is because either: (1) only one preference/event/action is provided and associated to a single speaker in the conversation, or (2) the preference/event/action of the two speakers are the same. Thus, the model only needs to understand ``\textit{what is said}'' in the conversation; the correct answer can be directly inferred purely from the text-based context.
For instance, in the dialogue in the middle of \autoref{fig:split}, only a single potential action is mentioned (e.g., the man wants to conquer his laziness), suggesting that the answer will only be ``do exercise''. In the other CBQ example, both speakers are disappointed by the slow service.

\subsection{Automatic Classification of ICQ and CBQ}
\label{sec:classification}
To automatically classify two-party SQA examples as ICQ or CBQ, we leverage a state-of-the-art language model (GPT-4) to assist the annotation, since previous work has shown that GPT-4 achieves human-level performance on simple tasks \cite{alizadeh2024opensourcellmstextannotation,törnberg2023chatgpt4outperformsexpertscrowd}.
Given the full transcript of an audio file with the associated speaker gender for each speaker turn,
we replace the mentioned speaker gender in the question with the opposite gender and add an additional answer option ``None of the above.'' Then, we request GPT-4 \cite{achiam2023gpt4} to answer the edited question based on the transcript with speaker gender tags.
The question is classified as follows:
\begin{itemize}
    \item If GPT-4 chooses the gender mentioned in the revised question, we define the question as ICQ. For the ICQ example in \autoref{fig:split}, the question will ask the man's preference and the answer will be fish.
    \item If GPT-4 answers ``None of the above,'' we define the example as CBQ since only one preference/action/event is associated with a single speaker. 
    \item If GPT-4 provides an the same answer as the answer to the original question, we classify it as CBQ since the asked preference/action/event must be associated with both speakers in the conversation.
\end{itemize}

\section{SQA Datasets}
\label{sec:datasets}
In this section, we present our analysis of CBQ and ICQ on Gaokao, an SQA dataset with real speech used to evaluate SpeechLLMs in~\cite{wavllm}. We find that a large portion of the speaker-related questions of this dataset is categorized as CBQ. 
In addition, we describe a simple synthetic SQA dataset, {\em What Do You Like?}, which we curate to allow a more controlled study of SpeechLLM performance. The full Gaokao dataset is available at WavLLM's github repository,\footnote{\url{https://github.com/microsoft/SpeechT5/tree/main/WavLLM}} and the data for our experiments can be accessed at \footnote{\url{https://github.com/wjk0925/slt2024-speechllm-speaker-understanding}}.

\subsection{Gaokao}
\label{sec:gaokao}
Gaokao contains 2000 spoken dialogs between a woman and a man, each comes with a multiple-choice question with three options. All questions use ``woman'' or ``man'' when referring to a speaker, thus by searching these two keywords, we select 919 speaker-related questions in the dataset for analysis. Example dialogues are in \autoref{fig:split}.

To classify these 919 SQA questions as ICQ or CBQ, we first obtain the full transcriptions of the dialogues using a Whisper-based speaker diarization + speech recognition system~\cite{whisper}\footnote{\url{https://github.com/MahmoudAshraf97/whisper-diarization}} and the inaSpeechSegmenter toolkit~\cite{ddoukhanicassp2018}, which has gender classification capability. The diarization and gender classification results are manually verified and corrected. We then follow the approaches described in \autoref{sec:issue_demo} and identify 773 CBQs and 146 ICQs.

\subsection{What Do You Like?}
Since questions in Gaokao require varying levels of reasoning and some dialogues contain potential gender hints such as names and titles (Mrs., Ms., Mr, ...), we curated a synthetic SQA dataset, {\em What Do You Like?} that follows the format of Gaokao (i.e. male-female dialogues, three-option multiple choice questions), but is much simplier to allow a more controllable analysis of SpeechLLMs' performance.

\begin{figure}[!h]
    \centering   
    \includegraphics[width=\linewidth]{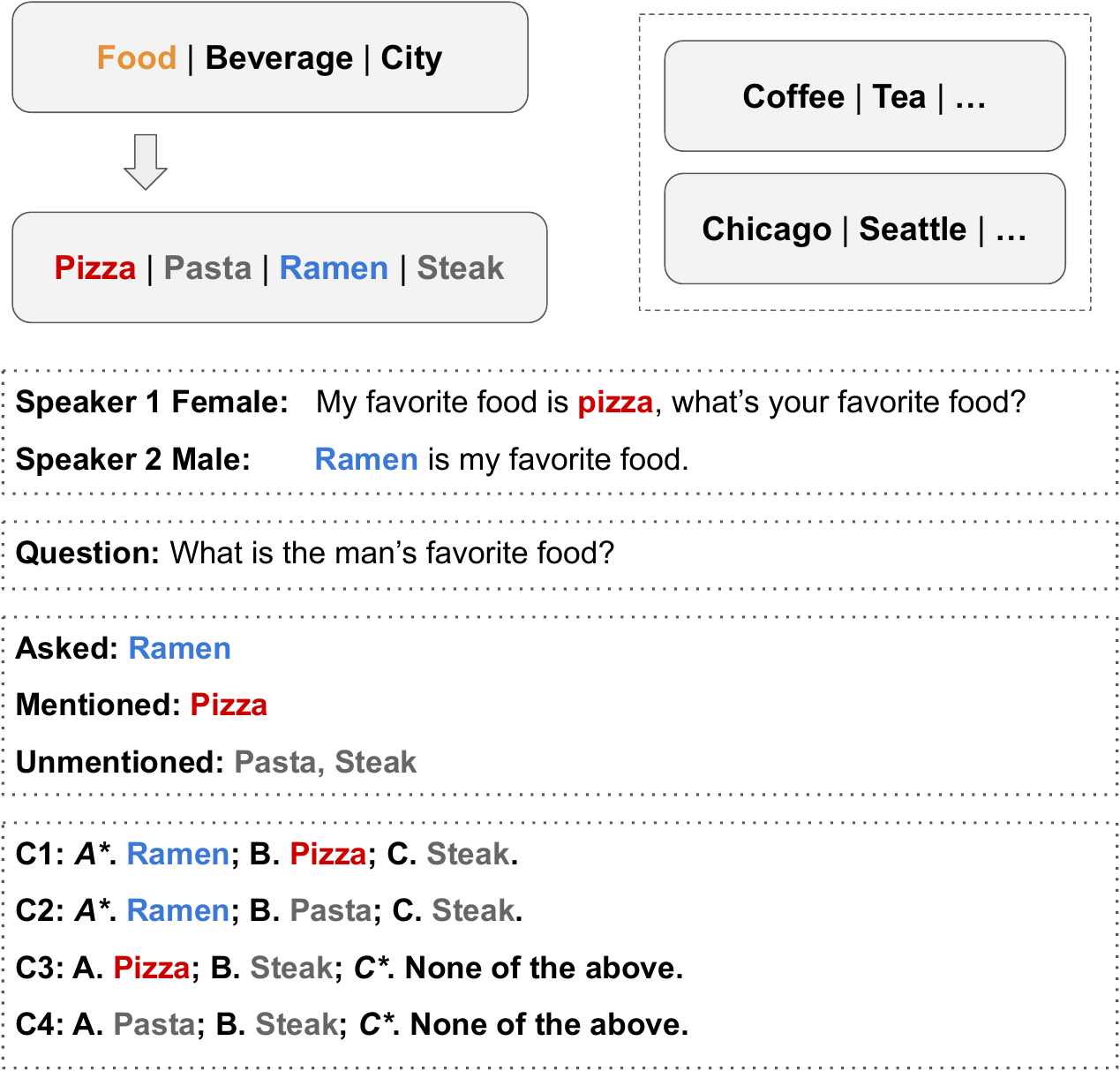}
    \caption{Example generation of a question and different answer options for {\em What Do You Like?}. The correct answer for each condition (C1:C4) is indicated by an asterisk.
    }
    \label{fig:Sample WDYL}
\end{figure}


The dataset contains 1000 single-turn spoken dialogues between a male speaker and a female speaker discussing their favorite \textbf{subjects} under a certain \textbf{topic}. As shown in Figure!\ref{fig:Sample WDYL}, we first generate the text dialogue by randomly selecting a \textbf{topic} and two \textbf{subjects}, both from a predetermined list of 3 topics (food, beverage, city) and 4 subjects for each topic. We then fill in them into one of 8 templates to make the text dialogue. Finally we use a text-to-speech model, StyleTTS-2~\cite{styletts2}, to convert the text dialogue into waveform. For TTS voices, we randomly pick a pair (one male and one female) from 5 different male voices and 5 different female voices in VCTK~\cite{vctk} and the train split of LibriTTS~\cite{libritts}, which StyleTTS-2 is trained on. 

We follow the Gaokao dataset and use multiple choice questions with three options for ``What Do You Like?''. For the question, we simply ask a random speaker's favorite subject using one of two templates: ``What is the \textbf{man/woman's} favorite \textbf{topic}?'' and ``Which \textbf{topic} does the \textbf{man/woman} like the most?''. We use man/woman to address the target speaker based on their randomly assigned gender during the TTS process. 
We generate four different sets of answer options according to the following conditions:
%

\textbf{C1:} (Asked, Mentioned, Unmentioned)

\textbf{C2:} (Asked, Unmentioned, Unmentioned)

\textbf{C3:} (Mentioned, Unmentioned, None)

\textbf{C4:} (Unmentioned, Unmentioned, None)

\noindent
where ``Asked'' is the favorite subject of the target speaker, ``Mentioned'' is the favorite subject of the other speaker, ``Unmentioned'' is another subject from the list, and ``None'' is the phrase ``None of the above.''


Note that the question is ICQ if asked in a open-end way, but conditions 2 and 4 change it into a CBQ. 
The dialogues contain no obvious gender hints and the questions only require simple information extraction to solve, making it possible to focus on SpeechLLMs' ability to identify speakers from voice differences under different answer conditions. 

\section{SpeechLLMs Performance Analysis}
\label{sec:results}

\subsection{Comparison Systems}
\subsubsection{SpeechLLMs}
\label{sec:speechllms}
We focus on WavLLM and Qwen-Audio as they are reported as the top SpeechLLMs for SQA tasks in \cite{wavllm, wang2024audiobenchuniversalbenchmarkaudio}. Qwen-Audio often gives free-form responses to multiple-choice questions instead of one option. To obtain the final answer, we prompt GPT-4 with the multiple-choice question and the Qwen-Audio’s response, and have it output one of ‘A’, ‘B’, ‘C’, or ‘X’, where ‘X’ means the response from Qwen-Audio does not correspond to any one of the choices A, B, or C.

\subsubsection{Text-only systems}
We implement two systems that can only use the ASR transcript by cascading ASR + LLM systems: Whisper-v3 + text-only WavLLM (denoted as WavLLM$\dag$), and Whisper-v3 + Llama-3-8B-Instruct \cite{whisper, Llama2} (denoted as Llama3). 
In addition, for oracle experiments on Gaokao, we evaluated Llama-3-8B-Instruct \cite{llama3modelcard} with speaker segmentation (denoted as 
Llama3+S),
and with both speaker and gender information (denoted as Llama3+SG), using the annotations generated in classifying ICQ and CBQ questions (see Section~\ref{sec:gaokao}).
The comparison of these different models provides an indication of the importance of gender information and the degree to which the speech models use it. The Llama3 systems also give free-form responses, we use the same approach as described in the previous section, prompting GPT-4 to obtain their final answers.


\subsection{Gaokao Results}
As reported in \cite{wavllm}, WavLLM reaches a 67.5\% accuracy and Qwen-Audio reaches a 54.2\% accuracy on Gaokao (full 2000 questions). In our study, we focus on the 919 speaker-related questions and show our results in \autoref{table:gaokao_r}.

\begin{table}[h!]
    \caption{The results on Gaokao measured in accuracy (\%).}
    \small
    \centering
    \begin{tabular}{ccccc}
    \toprule
    & Model & CBQ & ICQ & All \\ \midrule
    \multirow{2}{*}{SpeechLLMs} & WavLLM & 73.2 & 58.2 & 70.8  \\
    & Qwen-Audio & 62.2 & 43.2 & 59.2  \\ \midrule
    \multirow{4}{*}{LLMs} & WavLLM$\dag$ & 66.6 & 50.0 & 64.0   \\
    & Llama3 & 86.2 & 64.3 & 82.7   \\
    & Llama3+S & 86.3 & 66.4 & 83.1   \\
    & Llama3+SG & 88.7 & 91.8 & 89.2    \\
    \bottomrule
    \end{tabular}
    \label{table:gaokao_r}
\end{table}


As expected, when gender is not explicitly given, all models have significantly worse performance on ICQs than CBQs.  WavLLM gives somewhat better results than WavLLM$\dag$, but ICQ performance is not much better than chance. The Llama3 models all have better performance than the SpeechLLMs on CBQs, indicating the importance of a strong language model. The finding that the Llama models do better than chance on the ICQs is likely because of contextual indicators of gender. Comparing Llama3 to Llama3+S shows that adding speaker segmentation information has minimal benefit. The high ICQ score for Llama3+SG indicates the value of speaker information.


\subsection{Results on {\em What Do You Like?}}


The {\em What Do You Like?} dataset was designed to have four different answer set conditions to probe the behavior of SpeechLLM vs.\ text-only models. The conditions vary whether the ``Asked'' subject is in the answer set (if not, the correct answer is ``None'') and whether the ``Mentioned'' subject is in the set. The analyses below show the distribution of the three possible answers plus the non-compliant answer type ``X.'' Only non-oracle models are included; the Llama3+SG peformance is nearly perfect for all four conditions.

\begin{figure}[!h]
    \centering   
    \includegraphics[width=\linewidth]{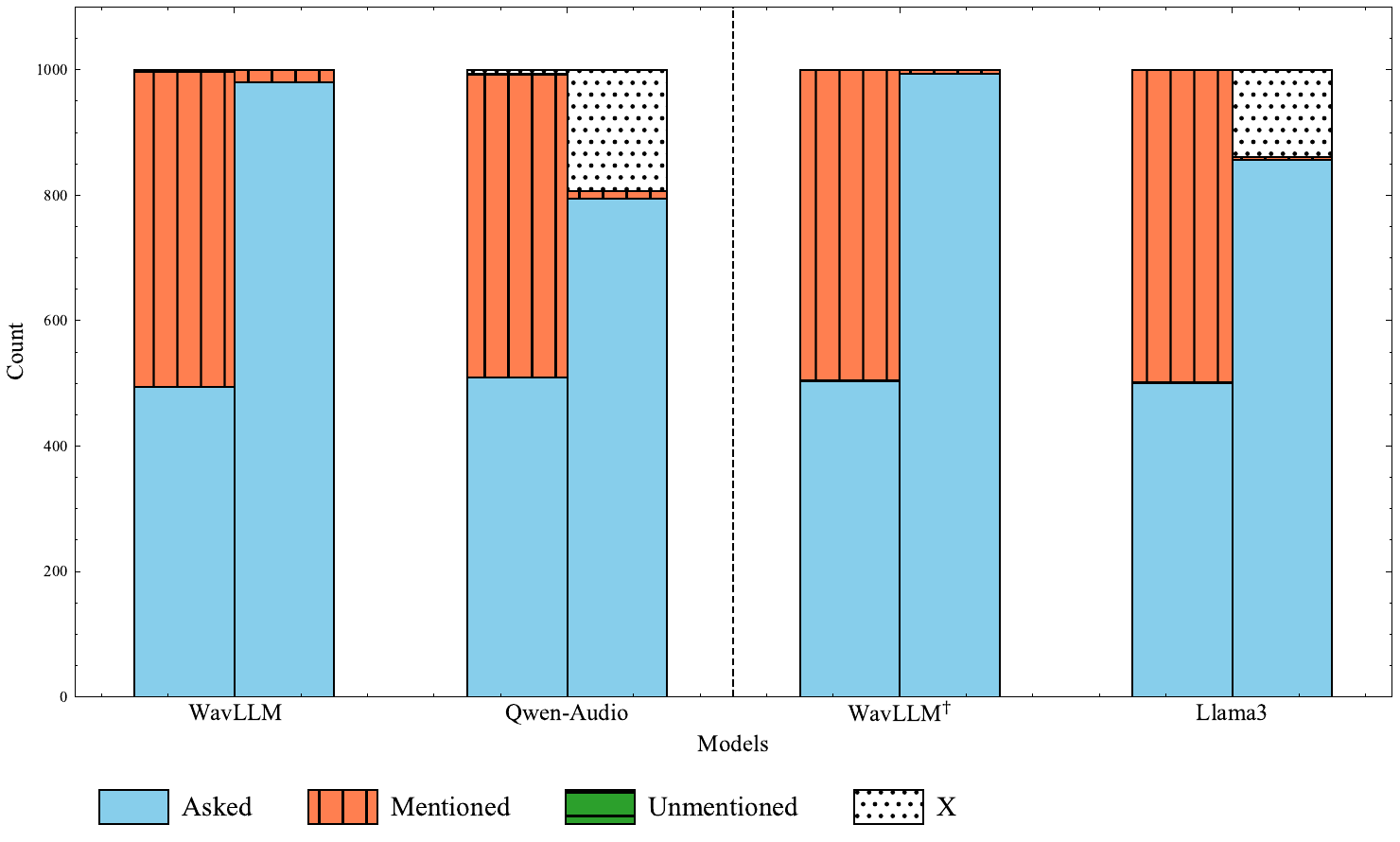}
    \caption{Number of times models choose each option for Condition 1 (left) and Condition 2 (right)}
    \label{fig:Resultes WDYL O12}
\end{figure}

In \autoref{fig:Resultes WDYL O12}, we compare the models' performance on Condition 1 (C1: \textbf{Asked, Mentioned, Unmentioned}) and Condition 2 (C2: \textbf{Asked, Unmentioned, Unmentioned}). As expected, all models perform at chance on C1. All have higher accuracy on C2, where there is no distracting Mentioned option. This indicates that SpeechLLMs are not using the voice information in the audio input to identify the asked speaker; instead they extract information from the dialogue content like LLMs. When there are two Unmentioned subjects in the answer set (C2), Qwen-Audio and Llama3 produce the \textbf{X} option fairly often.
It may be that the Unmentioned answers are sufficiently out of context that the models do not recognize this as a multiple choice task.

\begin{figure}[!h]
    \centering   
    \includegraphics[width=\linewidth]{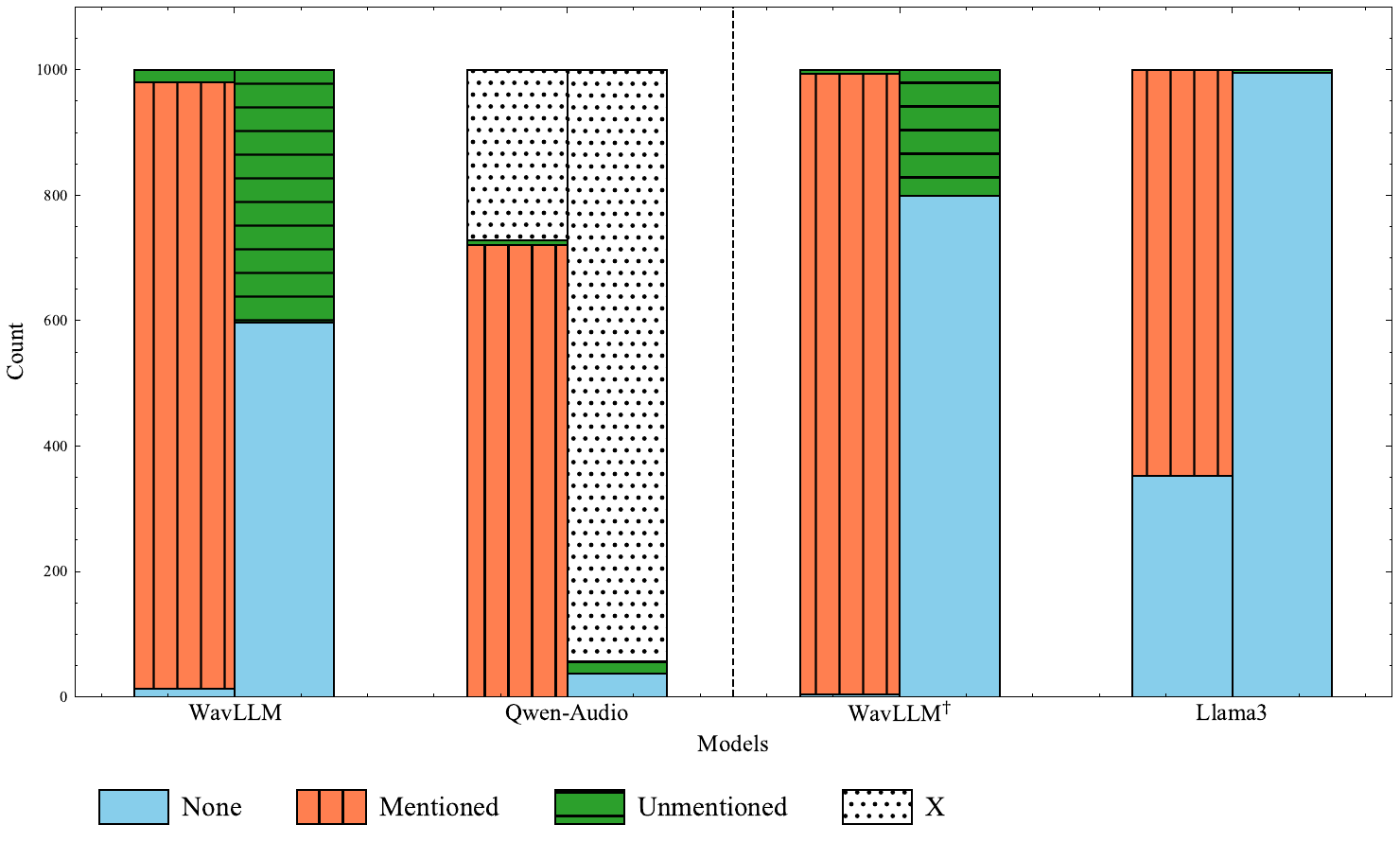}
    \caption{Number of times models choose each option for Condition 3 (left) and Condition 4 (right)}
    \label{fig:Resultes WDYL O34}
\end{figure}

\autoref{fig:Resultes WDYL O34} presents the models' performance on Condition 3 (C3: \textbf{Mentioned, Unmentioned, None}) and Condition 4 (C4: \textbf{Unmentioned, Unmentioned, None}). The correct answer in these cases is \textbf{None}, and the difference between C3 and C4 is again the presence of \textbf{Mentioned}. We observe that on C3, the two WavLLM models virtually always chose the Mentioned option, and the other models frequently choose it. Only Llama3 gets a sizable fraction correct. 
On C4, Qwen-Audio responses \textbf{X} for most of the questions, Llama3 reaches nearly perfect performance, and the two WavLLM models show a much stronger tendency to choose the correct answer \textbf{None}. 
These findings provide further support for the hypothesis that the SpeechLLMs are not using speaker voice information.

\section{Conclusion}
\label{sec:conclusion}
In this work, we categorize speaker-related spoken dialogue QA (SQA) questions into Identity-Critical Questions (ICQ) and Context-Based Questions (CBQ).
We hypothesize that current state-of-the-art speech large language models (SpeechLLMs) have limited capability in distinguishing speakers' voice characteristics and thus cannot solve ICQs as well as CBQs. To support our hypothesis, we propose an automatic classification approach for identifying ICQs and CBQs in large datasets, and conduct experiments and analysis using Qwen-Audio, WavLLM and text-only models on the Gaokao and {\em What Do You Like?} datasets. Our results show that WavLLM and Qwen-Audio obtain much lower accuracy on the speaker-related ICQ subset of Gaokao. Further, in the controlled {\em What Do You Like?} experiments, WavLLM performed similarly or worse than its text-only equivalent, supporting our hypothesis.

For future work towards improving SpeechLLM capabilities, it may be useful to explore alternative training techniques, e.g. pre-training with same/different speaker prediction tasks or RLHF with tasks that require speaker information. In addition, we argue that it is important to create a new SQA training and evaluation corpus that explicitly requires speaker identification capability and explores more attributes than gender.

\label{sec:conclusion}


\begin{thebibliography}{10}

\bibitem{qwenaudio}
Yunfei Chu, Jin Xu, Xiaohuan Zhou, Qian Yang, Shiliang Zhang, Zhijie Yan, Chang Zhou, and Jingren Zhou,
\newblock ``{Qwen-Audio}: Advancing universal audio understanding via unified large-scale audio-language models,''
\newblock {\em arXiv preprint arXiv:2311.07919}, 2023.

\bibitem{wavllm}
Shujie Hu, Long Zhou, Shujie Liu, Sanyuan Chen, Hongkun Hao, Jing Pan, Xunying Liu, Jinyu Li, Sunit Sivasankaran, Linquan Liu, and Furu Wei,
\newblock ``{WavLLM}: Towards robust and adaptive speech large language model,''
\newblock {\em arXiv preprint arXiv:2404.00656}, 2024.

\bibitem{gong2023ltuas}
Yuan Gong, Alexander~H Liu, Hongyin Luo, Leonid Karlinsky, and James Glass,
\newblock ``Joint audio and speech understanding,''
\newblock in {\em 2023 IEEE Automatic Speech Recognition and Understanding Workshop (ASRU)}. IEEE, 2023, pp. 1--8.

\bibitem{gong2023ltu}
Yuan Gong, Hongyin Luo, Alexander~H Liu, Leonid Karlinsky, and James Glass,
\newblock ``Listen, think, and understand,''
\newblock {\em arXiv preprint arXiv:2305.10790}, 2023.

\bibitem{whisper}
Alec Radford, Jong~Wook Kim, Tao Xu, Greg Brockman, Christine McLeavey, and Ilya Sutskever,
\newblock ``Robust speech recognition via large-scale weak supervision,''
\newblock {\em arXiv preprint arXiv:2212.04356}, 2022.

\bibitem{touvron2023llama1}
Hugo Touvron, Thibaut Lavril, Gautier Izacard, Xavier Martinet, Marie-Anne Lachaux, Timoth{\'e}e Lacroix, Baptiste Rozi{\`e}re, Naman Goyal, Eric Hambro, Faisal Azhar, et~al.,
\newblock ``{LLaMA}: Open and efficient foundation language models,''
\newblock {\em arXiv preprint arXiv:2302.13971}, 2023.

\bibitem{hu2021lora}
Edward~J Hu, Yelong Shen, Phillip Wallis, Zeyuan Allen-Zhu, Yuanzhi Li, Shean Wang, Lu~Wang, and Weizhu Chen,
\newblock ``{LoRA}: Low-rank adaptation of large language models,''
\newblock {\em arXiv preprint arXiv:2106.09685}, 2021.

\bibitem{bai2023qwentechnicalreport}
Jinze Bai, Shuai Bai, Yunfei Chu, Zeyu Cui, Kai Dang, Xiaodong Deng, Yang Fan, Wenbin Ge, Yu~Han, Fei Huang, Binyuan Hui, Luo Ji, Mei Li, Junyang Lin, Runji Lin, Dayiheng Liu, Gao Liu, Chengqiang Lu, Keming Lu, Jianxin Ma, Rui Men, Xingzhang Ren, Xuancheng Ren, Chuanqi Tan, Sinan Tan, Jianhong Tu, Peng Wang, Shijie Wang, Wei Wang, Shengguang Wu, Benfeng Xu, Jin Xu, An~Yang, Hao Yang, Jian Yang, Shusheng Yang, Yang Yao, Bowen Yu, Hongyi Yuan, Zheng Yuan, Jianwei Zhang, Xingxuan Zhang, Yichang Zhang, Zhenru Zhang, Chang Zhou, Jingren Zhou, Xiaohuan Zhou, and Tianhang Zhu,
\newblock ``Qwen technical report,''
\newblock {\em arXiv preprint arXiv:2309.16609}, 2023.

\bibitem{wavlm}
Sanyuan Chen, Chengyi Wang, Zhengyang Chen, Yu~Wu, Shujie Liu, Zhuo Chen, Jinyu Li, Naoyuki Kanda, Takuya Yoshioka, Xiong Xiao, Jian Wu, Long Zhou, Shuo Ren, Yanmin Qian, Yao Qian, Micheal Zeng, and Furu Wei,
\newblock ``Wavlm: Large-scale self-supervised pre-training for full stack speech processing,''
\newblock {\em IEEE Journal of Selected Topics in Signal Processing}, vol. 16, pp. 1505--1518, 2021.

\bibitem{superb}
Shu-Wen Yang, Po-Han Chi, Yung-Sung Chuang, Cheng-I Lai, Kushal Lakhotia, Yist~Y. Lin, Andy~T. Liu, Jiatong Shi, Xuankai Chang, Guan-Ting Lin, Tzu hsien Huang, Wei-Cheng Tseng, Ko~tik Lee, Da-Rong Liu, Zili Huang, Shuyan Dong, Shang-Wen Li, Shinji Watanabe, Abdel rahman Mohamed, and Hung yi~Lee,
\newblock ``{SUPERB: Speech processing Universal PERformance Benchmark},''
\newblock in {\em Interspeech}, 2021.

\bibitem{slue1}
Suwon Shon, Ankita Pasad, Felix Wu, Pablo Brusco, Yoav Artzi, Karen Livescu, and Kyu~J. Han,
\newblock ``Slue: New benchmark tasks for spoken language understanding evaluation on natural speech,''
\newblock in {\em ICASSP 2022 - 2022 IEEE International Conference on Acoustics, Speech and Signal Processing (ICASSP)}, 2022, pp. 7927--7931.

\bibitem{slue2}
Suwon Shon, Siddhant Arora, Chyi-Jiunn Lin, Ankita Pasad, Felix Wu, Roshan Sharma, Wei-Lun Wu, Hung-yi Lee, Karen Livescu, and Shinji Watanabe,
\newblock ``{SLUE} phase-2: A benchmark suite of diverse spoken language understanding tasks,''
\newblock in {\em Proceedings of the 61st Annual Meeting of the Association for Computational Linguistics (Volume 1: Long Papers)}, Anna Rogers, Jordan Boyd-Graber, and Naoaki Okazaki, Eds., Toronto, Canada, July 2023, pp. 8906--8937, Association for Computational Linguistics.

\bibitem{arora-etal-2024-evaluation}
Siddhant Arora, Ankita Pasad, Chung-Ming Chien, Jionghao Han, Roshan Sharma, Jee-weon Jung, Hira Dhamyal, William Chen, Suwon Shon, Hung-yi Lee, Karen Livescu, and Shinji Watanabe,
\newblock ``On the evaluation of speech foundation models for spoken language understanding,''
\newblock in {\em Findings of the Association for Computational Linguistics ACL 2024}, Lun-Wei Ku, Andre Martins, and Vivek Srikumar, Eds., Bangkok, Thailand and virtual meeting, Aug. 2024, pp. 11923--11938, Association for Computational Linguistics.

\bibitem{ao2024sdeval}
Junyi Ao, Yuancheng Wang, Xiaohai Tian, Dekun Chen, Jun Zhang, Lu~Lu, Yuxuan Wang, Haizhou Li, and Zhizheng Wu,
\newblock ``{SD-Eval}: A benchmark dataset for spoken dialogue understanding beyond words,''
\newblock {\em arXiv preprint arXiv:2406.13340}, 2024.

\bibitem{wang2024audiobenchuniversalbenchmarkaudio}
Bin Wang, Xunlong Zou, Geyu Lin, Shuo Sun, Zhuohan Liu, Wenyu Zhang, Zhengyuan Liu, AiTi Aw, and Nancy~F. Chen,
\newblock ``{AudioBench}: A universal benchmark for audio large language models,''
\newblock {\em arXiv preprint arXiv:2406.16020}, 2024.

\bibitem{salmonn}
Changli Tang, Wenyi Yu, Guangzhi Sun, Xianzhao Chen, Tian Tan, Wei Li, Lu~Lu, Zejun MA, and Chao Zhang,
\newblock ``{SALMONN}: Towards generic hearing abilities for large language models,''
\newblock in {\em The Twelfth International Conference on Learning Representations}, 2024.

\bibitem{alizadeh2024opensourcellmstextannotation}
Meysam Alizadeh, Maël Kubli, Zeynab Samei, Shirin Dehghani, Mohammadmasiha Zahedivafa, Juan~Diego Bermeo, Maria Korobeynikova, and Fabrizio Gilardi,
\newblock ``Open-source {LLMs} for text annotation: A practical guide for model setting and fine-tuning,''
\newblock {\em arXiv preprint arXiv:2307.02179}, 2024.

\bibitem{törnberg2023chatgpt4outperformsexpertscrowd}
Petter Törnberg,
\newblock ``{ChatGPT-4} outperforms experts and crowd workers in annotating political twitter messages with zero-shot learning,''
\newblock {\em arXiv preprint arXiv:2304.06588}, 2023.

\bibitem{achiam2023gpt4}
Josh Achiam, Steven Adler, Sandhini Agarwal, Lama Ahmad, Ilge Akkaya, Florencia~Leoni Aleman, Diogo Almeida, Janko Altenschmidt, Sam Altman, Shyamal Anadkat, et~al.,
\newblock ``{GPT-4} technical report,''
\newblock {\em arXiv preprint arXiv:2303.08774}, 2023.

\bibitem{ddoukhanicassp2018}
David Doukhan, Jean Carrive, Félicien Vallet, Anthony Larcher, and Sylvain Meignier,
\newblock ``An open-source speaker gender detection framework for monitoring gender equality,''
\newblock in {\em Acoustics Speech and Signal Processing (ICASSP), 2018 IEEE International Conference on}. IEEE, 2018.

\bibitem{styletts2}
Yinghao~Aaron Li, Cong Han, Vinay Raghavan, Gavin Mischler, and Nima Mesgarani,
\newblock ``{StyleTTS 2}: Towards human-level text-to-speech through style diffusion and adversarial training with large speech language models,''
\newblock in {\em Advances in Neural Information Processing Systems}, A.~Oh, T.~Naumann, A.~Globerson, K.~Saenko, M.~Hardt, and S.~Levine, Eds. 2023, vol.~36, pp. 19594--19621, Curran Associates, Inc.

\bibitem{vctk}
Junichi Yamagishi, Christophe Veaux, and Kirsten MacDonald,
\newblock ``{CSTR VCTK Corpus}: English multi-speaker corpus for {CSTR} voice cloning toolkit (version 0.92),''
\newblock 2019,
\newblock \url{https://datashare.ed.ac.uk/handle/10283/3443}.

\bibitem{libritts}
Heiga Zen, Viet Dang, Rob Clark, Yu~Zhang, Ron~J. Weiss, Ye~Jia, Zhifeng Chen, and Yonghui Wu,
\newblock ``{LibriTTS}: A corpus derived from librispeech for text-to-speech,''
\newblock {\em arXiv preprint arXiv:1904.02882}, 2019.

\bibitem{Llama2}
Hugo Touvron, Louis Martin, Kevin Stone, Peter Albert, Amjad Almahairi, Yasmine Babaei, Nikolay Bashlykov, Soumya Batra, Prajjwal Bhargava, Shruti Bhosale, Dan Bikel, Lukas Blecher, Cristian~Canton Ferrer, Moya Chen, Guillem Cucurull, David Esiobu, Jude Fernandes, Jeremy Fu, Wenyin Fu, Brian Fuller, Cynthia Gao, Vedanuj Goswami, Naman Goyal, Anthony Hartshorn, Saghar Hosseini, Rui Hou, Hakan Inan, Marcin Kardas, Viktor Kerkez, Madian Khabsa, Isabel Kloumann, Artem Korenev, Punit~Singh Koura, Marie-Anne Lachaux, Thibaut Lavril, Jenya Lee, Diana Liskovich, Yinghai Lu, Yuning Mao, Xavier Martinet, Todor Mihaylov, Pushkar Mishra, Igor Molybog, Yixin Nie, Andrew Poulton, Jeremy Reizenstein, Rashi Rungta, Kalyan Saladi, Alan Schelten, Ruan Silva, Eric~Michael Smith, Ranjan Subramanian, Xiaoqing~Ellen Tan, Binh Tang, Ross Taylor, Adina Williams, Jian~Xiang Kuan, Puxin Xu, Zheng Yan, Iliyan Zarov, Yuchen Zhang, Angela Fan, Melanie Kambadur, Sharan Narang, Aurelien Rodriguez, Robert Stojnic, Sergey Edunov, and Thomas
  Scialom,
\newblock ``Llama 2: Open foundation and fine-tuned chat models,''
\newblock {\em arXiv preprint arXiv:2307.09288}, 2023.

\bibitem{llama3modelcard}
AI@Meta,
\newblock ``Llama 3 model card,''
\newblock 2024,
\newblock \url{https://github.com/meta-llama/llama3/blob/main/MODEL_CARD.md}.

\end{thebibliography}

\end{document}